\documentclass[11pt]{article}

\usepackage{graphicx}
\usepackage{amsmath,amssymb} 
\usepackage{color}
\usepackage{geometry}
\usepackage{algorithm}
\usepackage{algpseudocode}
\usepackage{subcaption}

\begin{document}

\title{Multi-Scale Generalized Plane Match for Optical Flow} 

\author{Inchul Choi, Arunava Banerjee \\
\\
University of Florida\\
}

\maketitle

\begin{abstract}
Despite recent advances, estimating optical flow remains a
   challenging problem in the presence of illumination change, large 
   occlusions or fast movement. 
   In this paper, we propose a novel optical flow estimation framework 
   which can provide accurate dense correspondence and occlusion
   localization through a multi-scale generalized plane 
   matching approach.
   In our method, we regard the scene as a collection of 
   planes at multiple scales, and for each such plane, 
   compensate motion in consensus to improve match quality.
   We estimate the square patch plane distortion using
   a robust plane model detection method 
   and iteratively apply a plane matching scheme within
   a multi-scale framework.
   During the flow estimation process, our enhanced plane
    matching method also clearly localizes the occluded regions.
   In experiments on MPI-Sintel datasets, 
   our method robustly estimated 
   optical flow from given noisy correspondences, 
   and also revealed the occluded regions accurately.
   Compared to other state-of-the-art optical flow methods,
   our method shows accurate occlusion localization, comparable optical
   flow quality, and better thin object detection.
\end{abstract}

\section{Introduction}

Optical flow has been a focus of extensive research for several decades, and
has applications in diverse areas such as object tracking, motion estimation,
and robot navigation. Although the performance of optical flow algorithms have
incrementally improved over the years, the state-of-the-art tends to produce
large errors for videos with fast motion, illumination changes, and large
occlusions.

Computing an optical flow can be regarded as an estimation of dense
correspondences between two consecutive video frames. Recent trends in dense
matching algorithms  rely on sparse matches to handle large movement
problems\cite{Guney2017,Revaud2015,Revaud2015a,Yang2017}.  They use sparse
matching techniques for initialization and directly use interpolation
methods\cite{Hu2017,Revaud2015,Yang2017} to obtain dense correspondences. This
practice works well in many cases but has the shortcoming that it can generate
profoundly wrong answers if the sparse match initialization is erroneous.

\begin{figure}[t]
\begin{center}
   \includegraphics[width=\linewidth]{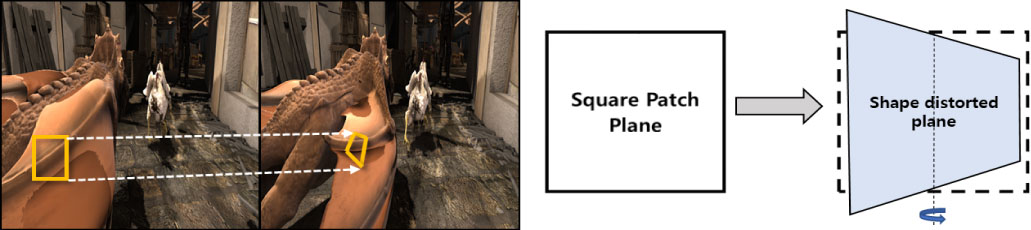}
\end{center}
   \caption{Patch Plane Distortion Problem. The square patch plane on the left-side is transformed to the shape distorted plane on the right-side after motion}
\label{fig:patchdistort}
\end{figure}

A contending approach for dense matching is the patch based method.  The patch
matching method aims to find the most comparable patch in a given image for
patches from a prescribed second image. This visual similarity between regions
in two images is the most important clue for large movement estimation. The
seminal work PatchMatch\cite{Barnes2009} algorithm provides visually adequate
dense correspondences called the nearest neighbor field (NNF).  Many dense
matching algorithms
\cite{Bailer2016,Bao2014,Bleyer2011,Chen2013,Fortun2016,Hornacek2014,Gadot2016,Heise2013,Li2015,Paper2016}
rely on, or are inspired by, the PatchMatch algorithm because of its efficiency
in computing the NNF.  However, the NNF is often too noisy to be directly
considered an optical flow because it lacks spatial regularization and in
addition, its patch comparison is based on a front-parallel assumption of
planes. Fig.\ref{fig:patchdistort} illustrates this intrinsic problem with the
patch based matching approach. When there is movement in the camera, or of
objects in the scene, between two consecutive images, square-shaped patches in
the first image correspond to distorted patches in the second. Since patch
match methods compare square regions with identical dimensions in the two
images, there is in principle no correct match, and this introduces errors. To
be a well-posed comparison scheme, one has to search through all square
distortion models for the best fit. However, searching through all such
distortions is computationally prohibitive.  In this paper, we resolve this
problem with a robust multi-scale plane detection method. Rather than search
through all distortions, we first determine a consensus distortion of an image
region from a given noisy NNF. This distortion is then compensated into the
patches for an improved match, and a resultant superior NNF.  It is well known
that in spite of NNFs being noisy they contain sufficient matching information
for robust plane motion estimation\cite{Chen2013}. Finally, making the
estimation multi-scale enables the detection of diverse size plane models that
exist in the scene. 

Of all the challenging conditions for dense matching, occlusion is probably the
most well identified that account for the lion's share of degradation of the
performance of optical flow algorithms. Occlusion, which by definition, has no
matching region in the second image, is detected indirectly by most methods via
a measure of the error of correspondence in the region. Cues such as high
photo-consistency error or forward-backward flow symmetry are used. However,
these cues are also applicable measures for evaluating flow quality in
non-occluded regions. This makes it difficult to distinguish a flow error in
the non-occluded region from invalid match errors in occluded regions, thereby
adversely affecting the flow quality in non-occluded regions.

In essence, detecting occlusion is a chicken-and-egg problem. If we have an
accurate optical flow, it can help localize the occlusion correctly, but
conversely, correct occlusion detection is necessary to obtain an accurate
optical flow estimation. Because of this intertwined relationship between
optical flow and occlusion, many algorithms have adopted a joint estimation
approach for occlusion and optical flow. We too take this joint approach during
optical flow estimation.

In this paper, we propose a novel framework that can estimate optical flow, and
localize occlusion, in a multi-scale patch plane based approach. Our framework
models the scene as a collection of multiple planes of different sizes and
incorporates non-front parallel plane compensation into the multi-scale patch
plane comparison method.  Our approach also estimates occlusion in a novel
multi-scale plane fitting framework. Occlusion arises from the motion of
distinct planes at depth discontinuities. Therefore, when we fit a homography
plane model to the local surface at the motion boundaries, occluded regions
naturally reveal themselves as regions that do not conform to any estimated
models in a photo-consistent manner.

To summarize, the main contributions of our work are as follows. We improve the
patch matching method by addressing the plane distortion problem. Our enhanced
plane matching scheme is combined with a multi-scale approach that robustly
detects most planar regions and occlusion in the scene. Experiments using our
framework shows that given a very noisy NNF, geometrically correct robust plane
matching can produce highly accurate dense matches. Although we use a simple
color consistency measure for patch plane comparison, the framework itself is
powerful enough to filter out most outliers and occluded regions.  Finally,
with robust interpolation, the estimated optical flow shows superior accuracy
in regions of the images with thin moving objects and comparable accuracy in
the remaining regions compared to state-of-the-art optical flow algorithms, as
well as accurate occlusion detection.

\section{Related Works}

Optical flow has been studied for decades and is used as a fundamental building
block in various vision systems. Here, we review recent work that relates to
patch match and occlusion estimation.

Patch based dense matching algorithms are frequently used in dense matching
methods. In a stereo problem\cite{Bleyer2011,Heise2013}, the PatchMatch
algorithm was used as the underlying matching scheme with additional smoothness
constraints. The paper modeled slanted planes to overcome the front-parallel
assumption and used the same propagation and random search policy as in
PatchMatch for dense match generation. Its plane model is however limited to
slants which means the degree of plane distortion is restricted to the stereo
case.

In optical flow algorithms, \cite{Li2017,Paper2016} used PatchMatch in a
coarse-to-fine framework to obtain reliable matches for large displacement of
objects and used edge preserving interpolation methods\cite{Revaud2015} to
obtain dense matching. The Patch based belief propagation
method\cite{Hornacek2014,Li2015} and edge-preserving PatchMatch
algorithms\cite{Bao2014} are other examples of PatchMatch inspired approaches.
However, none of these have attempted to overcome the front-parallel plane
assumption.

The occlusion problem is also treated in many optical flow algorithms and their
approach for handling occlusion can be categorized into two basic themes.
The more common approach is to filter out occluded pixels as outliers during
estimation. The filtering out can decrease the adverse effects of occlusion to
non-occluded regions during flow
estimation\cite{Butler2012,Chen2016,Guney2017,Li2015,Paper2016}. The
constraints for outlier filtering are the usual cues of occlusion detection,
such as high-intensity error or forward-backward flow
consistency\cite{Ince2008}. These approaches rely on a robust data term to
minimize the effect of outliers, and the filtered outlier regions are filled in
during a post-processing step. These techniques are generally initialized with
reliable sparse matching methods and they perform interpolation with the given
reliable matches to expand the sparse matches to dense flows.

The other approach to handling occlusion is a joint estimation of occlusion and
optical flow. This approach explicitly includes an occlusion variable into its
optimization objective\cite{Sun,Xiao2006}. They assign an occlusion label to
each pixel based on a high photo-consistency violation or flow-symmetry
constraint, and assign a constant penalty as a separate term. However, these
methods also cannot distinguish error due to occlusion from error due to
poor matches.  Therefore, they sometimes  arrive at incorrect occlusion
localization when the scene has large occluded regions.

Like other approaches, we estimate occlusion and optical flow based on high
photo-consistency error and forward-backward flow symmetry. However, different
from the above methods, we detect occluded regions based on patch plane
transform models. A region that does not have any plane transform similar to
its neighboring planes, in addition to showing high color-consistency error and
forward-backward plane projection asymmetry, is labeled an occluded region.

\begin{figure*}
\includegraphics[width=\linewidth]{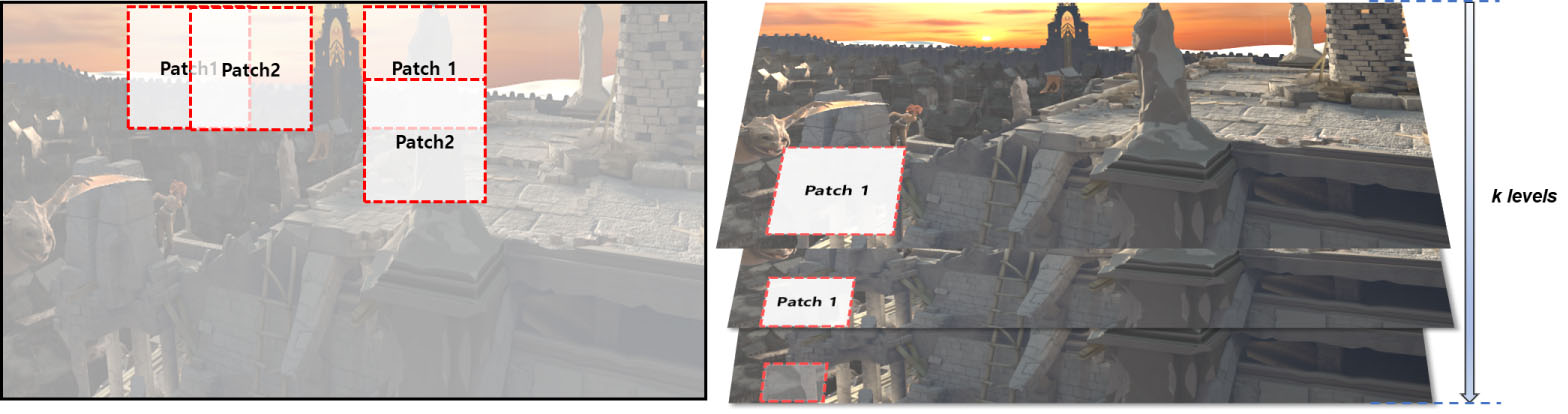}
\caption{Multi-Scale Generalized Plane Match (MSGPM). Figure shows how patch window overlaps in each level and  in the multi-scale(level) with different size}
\label{multiscale}
\end{figure*}

\section{Multi-Scale Generalized Plane Match}

We now present our robust multi-scale generalized plane matching framework
(MSGPM) and discuss its core features. Our method builds robust plane matches
to refine the optical flow, and operates in a top-down fashion in a multi-scale
architecture.  We first detail the robust patch plane match at a single level
in Section 3.1, following which we describe the multi-scale structure as well
as the propagation of the refinement between levels in Section 3.2 and 3.3. Occlusion
detection and the final filling in scheme is presented in Section 3.4.

\subsection{Robust Plane Match}

Our baseline dense matching is generated by the PatchMatch
algorithm\cite{Barnes2009}. It provides a dense correspondence of pixels from
the two images, called NNF, which although reasonably good isn't accurate
enough to be regarded as an optical flow.  The primary reason for its match
error is due to occlusion and the front-parallel plane assumption. Most patch
based matching methods rely on square shaped identical patches to compare
regions from two images.  However, in many cases, a square shaped patch plane
in one image is distorted due to motion of the camera or of objects;
particularly, in the case of zoom in-out or non-rigid motion of objects.
Strictly speaking, an exactly matching square patch may not exist in the second
image. The obvious solution of enumerating and comparing against all possible
plane distortion models for each patch comparison is computationally
prohibitive. Instead, we resolve this problem by first detecting a consensus
plane distortion model from the given noisy NNF and then compensating that
plane distortion to improve match quality.

\subsubsection{Homography Plane Model}
We assume that most regions of a scene can be approximated by a piecewise
homography plane model. For two consecutive video frames, i.e., images
$I_{1}$,$I_{2}$ , each matching pair on the same local plane
$\mathcal{M}_{1}=(p_{1},p_{1}')$, $p_{1} \in I_{1}$ , $p_{1}' \in I_{2}$,
satisfies both the homography transform $H_{p_{1}'\rightarrow p_{1}}$ and
a color-consistency (intensity consistency) assumption,

\begin{equation}
p_{1} = H_{p_{1}'\rightarrow p_{1}}p_{1}'.
\label{eq1}
\end{equation}

\begin{equation}
I_{1}(p_{1}) - I_{2}(p_{1}') = 0.
\label{eq2}
\end{equation}
For the point pairs that satisfy the above relations, the corresponding local plane model $H_{p_{i}\rightarrow p_{i}'}$ can be assigned. In our framework, we used a least square method to obtain a homography model from 4 correspondences. 

\subsubsection{Robust Model Estimation}

Instead of estimating a plane model directly from noisy matches in a pair of
patches from the two images, we use a Random Sample Consensus (RANSAC) based
approach\cite{Chum2008} for robust detection of a model. The RANSAC method
robustly detects a dominant model by finding the largest inlier support model
from a random sampling of input matches.  The advantage of using RANSAC is not
only its robustness to noise but also its capacity to detect a single dominant
model in the patch region. Therefore, even when there exists occlusion in the
patch, the technique can detect a plane corresponding to the non-occluded
region correctly. 

To account for cases where there exist more than one plane in the patch region,
we use overlapping patches.  We overlap the patch window with the previous
patch region using half ratio as shown in Fig.\ref{multiscale}. This overlapping
between two consecutive patch regions enables detection of multiple planes in a
local region and also allows for smooth transitions between planes.

\subsubsection{Plane Distortion Compensation}

The detected homography plane model is used to compensate for the corresponding
plane motion to improve the match quality in the local region.  Our homography
plane model choice and subsequent distortion compensation method is based on
a classical optical flow estimation principle. It warps the second image toward
the first image using the detected homography plane transform (Eq\ref{eq1}) and
evaluates whether the color-consistency (Eq\ref{eq2}) is satisfied within the
patch regions as follows.

Assume that a homography plane model $H_{1 \rightarrow 2}$ is detected for a
given patch region $R_{i}$. Then, the plane motion compensation is performed by
warping $I_{2}$ toward $I_{1}$ by the $H_{2 \rightarrow 1}$ induced flow. For
each pixel $p_{i}$ in the region $R_{i}$ of $I_{1}$, $p_{i} \in R_{i}$, there
exist a corresponding pixel $p_{i}'$ of the warped second image $I_{H_{2
\rightarrow 1}}$ which matches against the position of $p_{i}$. The
color-consistency error is then

\begin{equation}
 \delta_{p_{i}} =  I_{1}(p_{i}) - I_{H_{2 \rightarrow 1}}(p_{i}'),  p_{i} \in R_{i}.
\label{eq3}
\end{equation}

The color-consistency violation $\delta_{i}$ represents the color difference
between $I_{1}$ and warped image  $I_{H_{2 \rightarrow 1}}$ at the pixel
location $p_{i}$. For each $p_{i}$, the corresponding pixel $p_{i}'$ should
satisfy the color-consistency relation (Eq\ref{eq2}) if it is on the same plane
$H_{i}$. However, since the plane in the scene is not strictly flat and there
also exists noise and fitting error, we use an error threshold $\epsilon$ as a
flexibility parameter for how well a pixel conforms with a given plane model. 

For a pixel $p_{i}$ , $p_{i} \in R_{i} $, when the color-consistency error is
smaller than the threshold $\epsilon$, the pixel is regarded as an inlier of
the estimated plane $H_{i}$,

\begin{equation}
L_{H_{i}}(p_{i}) =
\begin{cases}
|\delta_{p_{i}}|  &\text{if $|\delta_{p_{i}}| < \epsilon$}\\
\epsilon  &\text{otherwise.}
\end{cases},
 p_{i} \in R_{i}.
\label{eq4}
\end{equation}

The robust loss $L_{H_{i}}(p_{i})$ represents how poorly a point $p_{i}$
conforms with the homography plane model $H_{i}$ in the region $R_{i}$.  If the
color-consistency error is larger than $\epsilon$, then the point $p_{i}$ does
not get assigned to the $H_{i}$ plane.  For a point in the patch overlapping
region, if it is detected as an inlier for two different planes, the plane with
the smaller $L_{H_{i}}(p_{i})$ is assigned to the point.

The patch plane motion compensation identifies a region that conforms with the
estimated plane model in the patch. We assign the homography induced flow to
the detected region as the refined flow. As a result of this procedure, even
though the initial matches are very noisy, if the estimated plane model matches
any planar part of the patch region, we can obtain an enhanced correspondence
for that region.

The one problem with this approach is that the plane detection is affected by
the size and position of the patch. The patch size limits the detectable plane
sizes and the patch location limits the detectable plane positions in the
scene.  Scenes generally consist of a diversity of sizes of planes and when a
plane size is smaller than the dominant plane that exists in the patch window,
it is likely to be neglected by RANSAC since RANSAC only finds a single
dominant plane.  We address this problem by applying a residual-region based
model detection and a multi-scale extension to the above robust plane matching
method.

\begin{figure}[t]
\begin{center}
\includegraphics[width=0.7\linewidth]{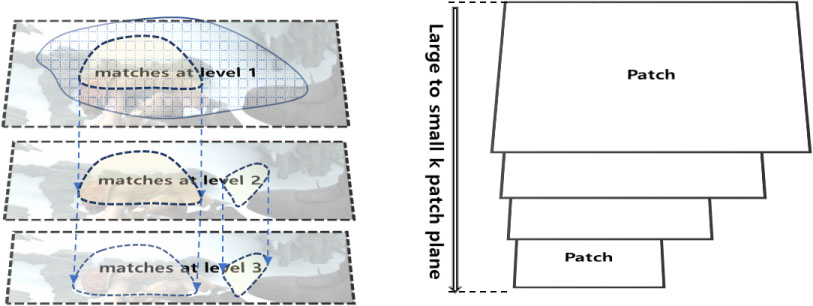}
\end{center}
\caption{From current level, the region of reliable correspondences are propagated to the next level.}
\label{fig:reliablematch}
\end{figure}

\subsection{Residual Region based Model Detection}

In the patch window based RANSAC model detection scheme, if more than one plane
model exists in the patch window region, the small plane (based on the size of
the region it encompasses) can be overlooked even when the patch windows are
overlapped. This is because RANSAC always chooses the most dominant plane from
the input matches, and therefore, the detectable planes are limited by the size
and positions of the patch window. To address this problem, we detect planes in
two stages.  In the first stage, we detect and assign a plane by moving a fixed
size patch window with overlaps in a scan-order. Following this, there might
remain residual regions which do not have any assigned homography model. In the
second stage, we only use the residual regions in a patch window as the input
to the RANSAC based model detection. Accordingly, any detected plane from the
region is assigned only to the residual region using the color-consistency
measure.  This residual region based model detection enables us to detect as
many planes as possible with the same fixed size patch window. For smaller
planes that may exist in the same local region, we resort to detection using a
Multi-Scale patch window based approach as described next.

\subsection{Multi-Scale Plane Match}

A single scale patch window based robust plane matching scheme detects a
dominant plane model per patch region. To discover the diverse sizes of planes
that exist in the scene, we need diverse sizes of patch windows.  Our
multi-scale patch plane approach applies a large to small patch size variation
scheme within a multi-level architecture. We begin with the largest patch
window size, the first level, and the robust plane match is applied to the
entire image.  At the next level, we repeat the same procedure with a slightly
smaller patch window size. We decrease the patch window size gradually until we
reach the bottom level, corresponding to the finest scale. Fig.\ref{multiscale}
illustrates this multi-scale structure.

For a given level $l \in \{ 1,...,k \} $ and maximum patch window radius
$w_{max}$, each level uses window radius $w_{l} = w_{max} - (l-1)* \Delta w$,
where $\Delta w$ is the window radius decrement per level transition.

\begin{figure*}
\includegraphics[width=\linewidth]{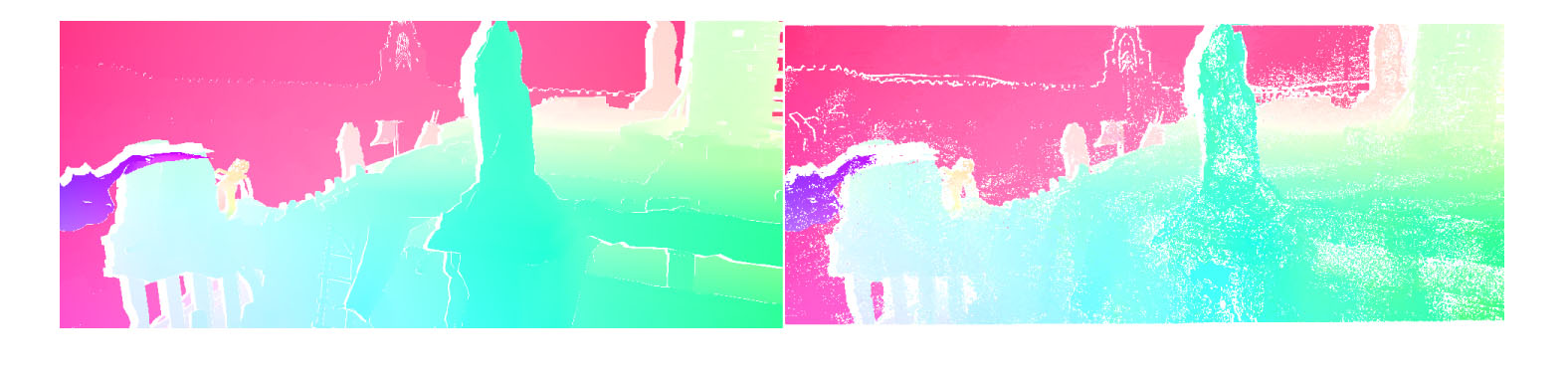}
\caption{Multi-Scale Generalized Plane Match (MSGPM) result (flow color scheme of the benchmark is used). Left: ground truth optical flow with occluded regions removed. Right: MSGPM match }
\label{fig3}
\end{figure*}

Collecting the many levels of this multi-scale framework, we obtain a full set
of plane fit flows. These plane flows from each level $l$ vertically overlap
each other for each location of the image. These multi-level flows are merged
using the same method that is used for overlapping patch regions (Eq\ref{eq3}):
for each pixel location $p_{i}$ in the image for which a flow has to be
assigned, the plane model with the minimum color consistency error
$L_{H_{i}}(p_{i})$ is assigned as the winner.  The only difference is that when
merging flows from different levels, we give higher priority to lower levels
with larger window sizes since as the window size decreases, the effect of
noisy matches and the concomitant flow distortions rise.  The result at the end
of this stage, i.e., multi-scale with robust plane detection, is shown in
Fig.\ref{fig3} as a color-coded map. When compared to the ground truth optical
flow in Fig.\ref{fig3}, our method shows highly reliable dense matches in the
non-occluded regions. 

\subsubsection{Propagation of Reliable Models between Levels}

When we transition from one level to the next, we reflect the reliable
estimation result to the next level's initial match via the NNF. As illustrated
in Fig.\ref{fig:reliablematch}, for the regions with low color-consistency
error with assigned homography model, the correspondences are propagated to the
initial matches for the next level. This propagation procedure increases the
reliability of matches by decreasing noisy outlier correspondences.
Algorithm.\ref{planematch} gives an overview of the one level plane matching
method in the multi-scale framework.

\begin{algorithm}
   \caption{Robust Generalized Plane Match}
    \begin{algorithmic}[1]      
    \State For each level $l$  with patch size $w_{l}$ 
    \State $l\in \{ 1,..,k \}$
    \State /* First Stage */
        \For{ each NNF  $patch_{i}$ of size $w_{l}$}
        	\State /* robust model estimation */
        	\If{ plane model $H_{i}$ is detected }
        	   
        	   \State warp $I_{2}$ by $H_i$ induced flow
        	   \State check color-consistency $L_{H_{i}}$
        	   \State detect consistent region $r_{plane}$
        	   \If{ plane symmetry $>$ $\eta$ }
        	   
        	     \State assign flow to $r_{plane}$
        	    
        	   \EndIf
        	   
            \EndIf         
        \EndFor
        \State
        \State /* Second Stage */
        \For{ each NNF $patch_{i}$ with residual region $Res_{i}$}
        	\State robust model estimation and assignment w.r.t $Res_{i}$
        	
        \EndFor
        \State
		\State $w_{l} = w_{max} - (l-1)*\delta w$
    	\State update NNF with reliable matches	for the next level
    	\State Run PatchMatch for the unassigned regions of level $l$
\end{algorithmic}
\label{planematch}
\end{algorithm}

\subsubsection{Match Refinement for Unassigned Regions}

Each level of MSGPM identifies both multiple planes that exist in the scene as
well as occluded regions. However, after the two stages of plane detection,
there might remain regions that are not occluded and are yet un-detected (such
as when there are objects with large non-rigid motion). To further identify the
flow for such regions at the next level, we run PatchMatch restricted to the
unassigned regions of the current level's flow. The regions already assigned to
homography models are masked and used for providing possible match candidates
for the random search in PatchMatch. This restriction to unassigned regions
decreases the search space and increases the probability of finding correct
matches from the NNF.

\begin{figure}
\centering
\includegraphics[width=\linewidth, height=6cm]{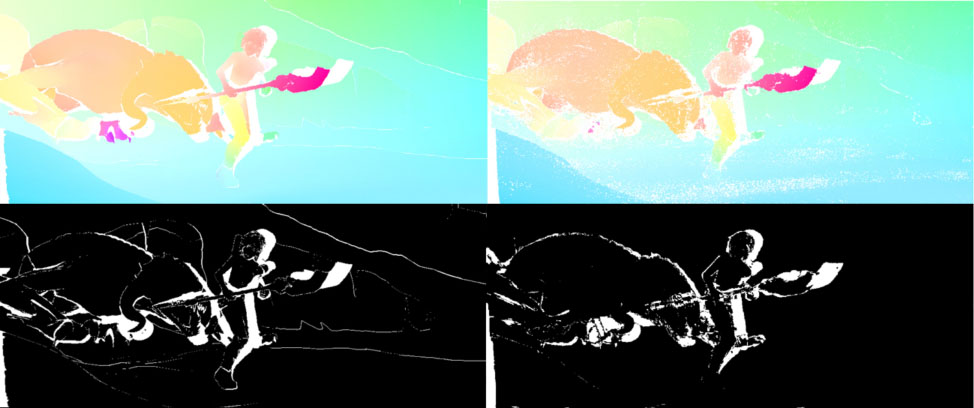}
\includegraphics[width=\linewidth, height=6cm]{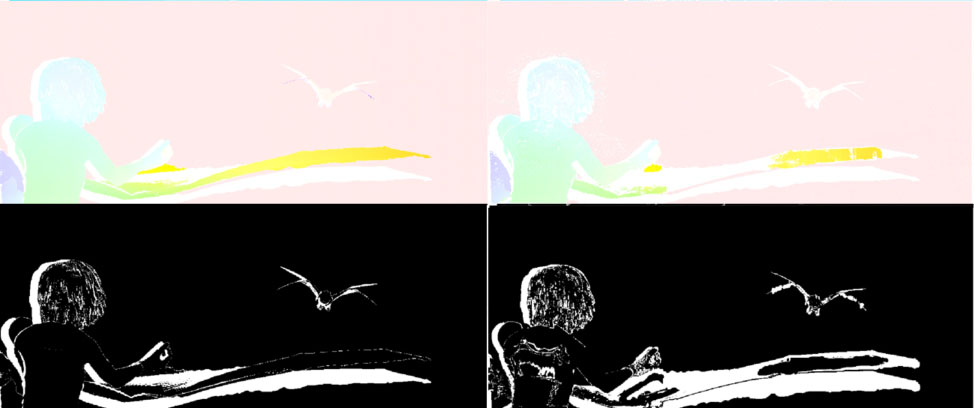}
\includegraphics[width=\linewidth, height=6cm]{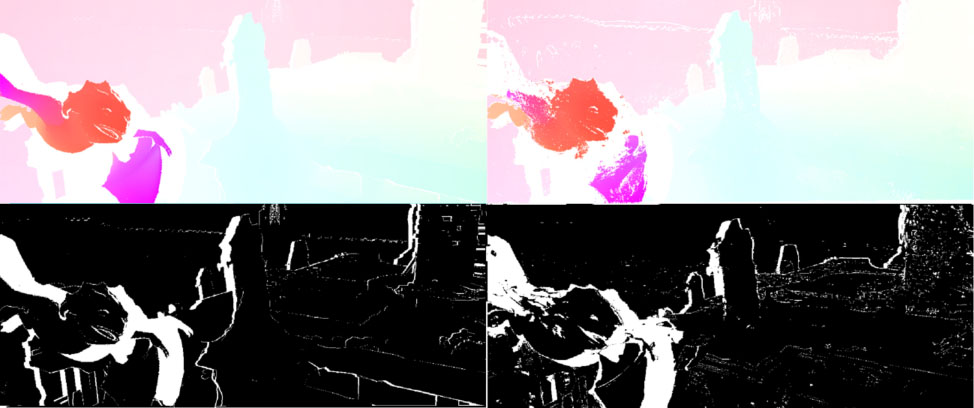}
\caption{MSGPM generated correspondences (color coded) and occlusion estimation. (for each image set) upper row left: Ground truth optical flow of non-occluded region, upper row right: MSGPM result, lower row left: Ground truth occlusion map ,lower row right: MSGPM occlusion estimation result.}
\label{occestm}
\end{figure}
\begin{figure}
\centering
\includegraphics[width=\linewidth, height=6cm]{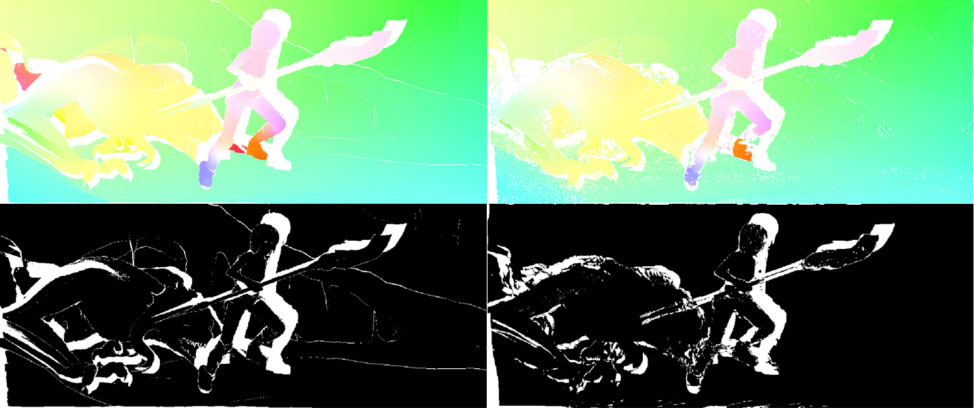}
\caption{MSGPM generated correspondences (color coded) and occlusion estimation. (for each image set) upper row left: Ground truth optical flow of non-occluded region, upper row right: MSGPM result, lower row left: Ground truth occlusion map ,lower row right: MSGPM occlusion estimation result.}
\label{occestm2}
\end{figure}

\subsection{Occlusion Detection and Interpolation}

Since, by definition, for an occluded region in a video frame, there does not
exist a corresponding region in the second frame and consequently no homography
match, in an ideal world, our multi-scale generalized plane matching method
would detect and assign planes to all regions of a scene except for occluded
regions. In experiments, we have found that this is indeed the case for most
examples and that occluded regions remain empty (i.e., devoid of an assigned
flow) after the completion of the overall flow estimation procedure.
However, because of accidental matches in the NNF of parts of occluded
regions to color consistent yet erroneous parts of the second image, there are
cases where parts of the occluded regions get incorrectly assigned planes.
These noisy matches can infiltrate the plane model detection procedure when the
patch size is small and close to the size of the erroneous region.  We identify
these incorrectly matched occluded regions using the following criteria.

\begin{itemize}
\item Forward-backward plane projection symmetry.
\item Many-to-one matched planar region with higher color-consistency error.
\item Overlap ratio between inliers from color-consistent region and inliers from the RANSAC method.
\end{itemize}

The first occlusion cue is frequently used in other algorithms. In our case,
when we detect a plane model from the patch window, we also estimate the
backward plane model from the backward NNF. The input matches are the backward
NNF in the projected region of the estimated forward plane model. If the
detected forward plane region is not overlapping with the backward plane's
projected region, the detected region is defined to be invalid. We set the
symmetric region overlapping threshold as $\eta$ and discard a plane model that
is below this threshold.

The second cue is similar to the match uniqueness criteria in stereo matching:
most of the non-occluded regions in an image are uniquely matched to regions in
the second image. When some planar region in the first image is mapped to an
already matched region in the second image, then among those multiple matched
candidate regions, the one with the higher color-consistency error is likely to
be an occluded region.

The last cue tests for agreement between the RANSAC model estimation and the
color-consistency based inlier detection scheme. The plane model estimated from
the matches should agree with the color-consistent planar region for the plane
matching scheme to be correct. If this is violated we deem the match incorrect.

Based on above three criteria, we detect incorrectly matched regions from each
level. When merging final flows, occluded regions are deemed to be regions with
no correct plane model.

\subsubsection{Filling and Occlusion Map}

From the obtained reliable correspondences, we fill as much of the unmatched
regions as possible using homography model propagation.
The homography model propagation is a method that propagates the best detected models 
based on color-consistency cost to neighboring unfilled pixels.

Finally, we use an edge-preserving interpolation method
(Epic-flow)\cite{Revaud2015} with its default parameters to fill in all remaining
regions. The flow interpolated by Epic is merged with the MSGPM flow
based on a color-consistency cost. 

Given the final interpolated flow, we compare the color-consistency error of
the two images by warping and then detect occluded regions as those above an
error threshold.
Results from our Multi-Scale Generalized Plane Match (MSGPM) and its
corresponding occlusion estimation maps are shown in Fig.\ref{occestm} and Fig.\ref{occestm2}.  As is
clear, the results are very close to the ground truth in both quality and
occlusion detection.

\section{Experiments}

We evaluated our framework on the MPI-Sintel\cite{Butler2012} dataset for
optical flow. MPI-Sintel is a challenging benchmark based on an animated movie
that contains several large movements of objects, large motion and zoom-in/out
of the camera, large occlusions, and illumination changes. We compared our
technique, MSGPM, against several state-of-the art methods over 
difficult segments of the video that corresponded to large occlusion and large
movement of thin objects. Experiments were run over the entire dataset. In
general, it is well known that all techniques do well on easier frames, and we
found this to be true in our experiments.

In all experiments, an NNF resulting from the running of the PatchMatch
algorithm in $7\times7$ patch size was used as the initial input to the MSGPM
algorithm. Furthermore, MSGPM used 2 levels in the multi-scale framework with
the maximum patch window radius set to $w_{max}=40$ and the window radius
decrement per transition level set at 20 ($\Delta \omega = 20$).  At the end of
each level, the plane projected flow based forward-backward symmetry was
checked to filter out outliers and refine the occluded region estimation. 

In MSGPM, when the number of levels, $l$, was increased, the quality of the
flow did increase. However, the improvement was not significant enough to
warrant the over-head introduced by adding more levels. We found that $l=2$ was
the minimum number of levels with non-significant performance degradation. We
therefore set $l=2$ in all our experiments. 

Comparisons between the algorithms was based on the mean endpoint error(EPE).

\begin{figure}
\centering
\includegraphics[width=\linewidth, height=6cm]{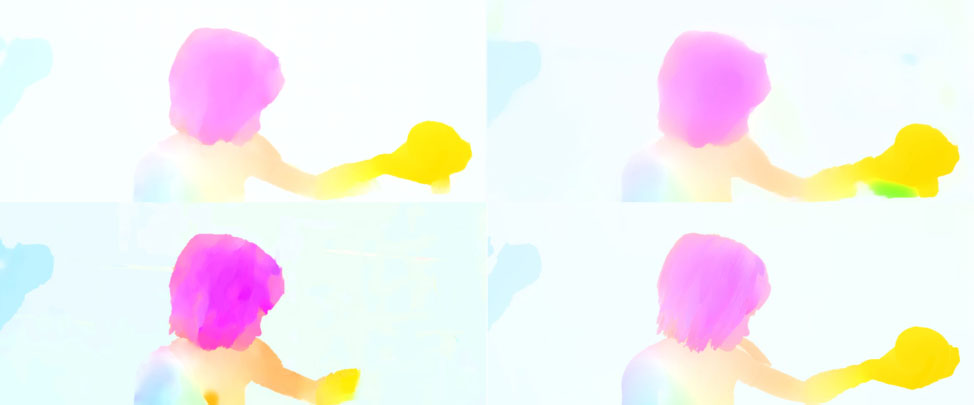}
\includegraphics[width=\linewidth, height=6cm]{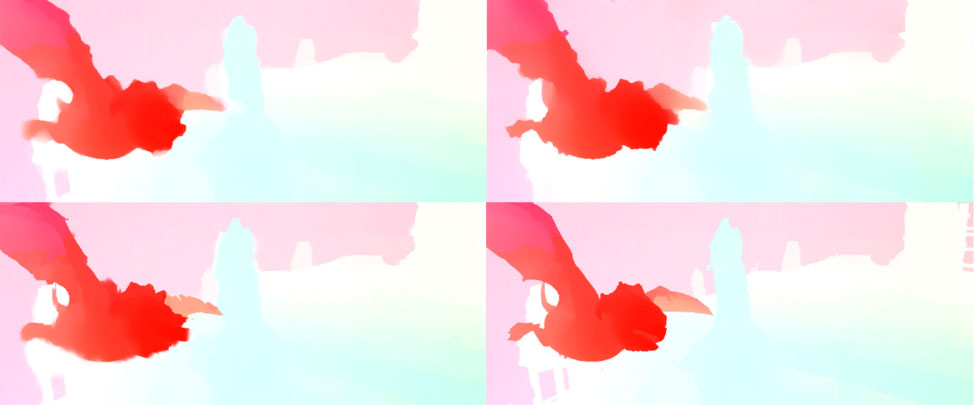}
\includegraphics[width=\linewidth, height=6cm]{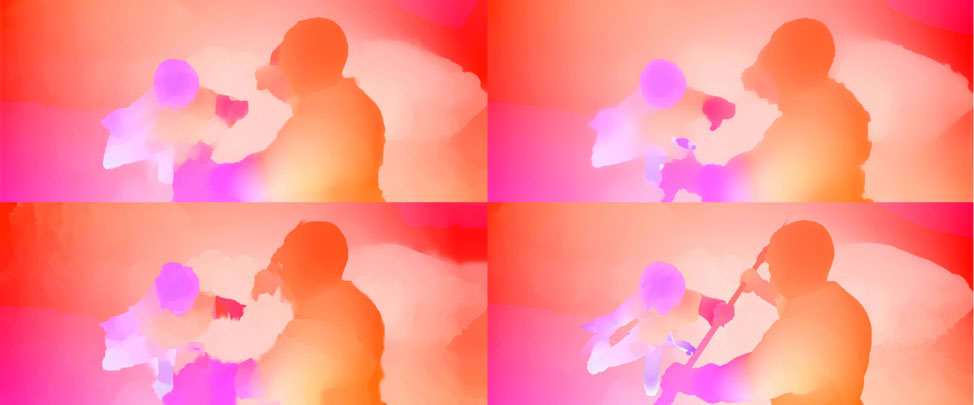}
\caption{Color coded optical flow results.(for each image set) upper row left: MSGPM+Epic, upper row right:  CPM2-flow, lower row left: SPMBP-flow, lower row right: Ground truth optical flow}
\label{interoplatedflow}
\end{figure}


\begin{table}[t]
\begin{center}
\begin{tabular}{p{3cm} p{2cm} p{2cm} p{2cm}}
\hline
Method & EPE nocc. & EPE occ. & EPE all\\
\hline
MSGPM+Epic                  & 3.319  & 19.368  & 4.488   \\
CPM2-flow\cite{Paper2016}   & 0.796  & 20.171  & 2.208  \\
EPIC-flow\cite{Revaud2015}  & 1.036  & 18.207  & 2.287  \\
SPMBP\cite{Li2015}          & 1.785  & 24.881  & 3.468  \\
\hline
\end{tabular}

\caption{EPE (Endpoint Error) results on MPI-Sintel dataset.}\label{tab:aee}
\end{center}
\end{table}

The evaluations were performed on the final interpolated results. As noted
earlier, MSGPM+Epic refers to interpolation by Epic \cite{Revaud2015}. Table \ref{tab:aee}
demonstrates that our method shows comparable performance with other
state-of-the-art optical flow algorithms. Fig.\ref{interoplatedflow} shows the
computed optical flow of each algorithm for an exemplar image pair. The optical
flow is represented by the color code set by the benchmark. As is clear,
compared to other state-of-the-art methods, MSGPM generates optical flow of
similar quality.


\begin{figure}
\centering
\includegraphics[width=\linewidth, height=6cm]{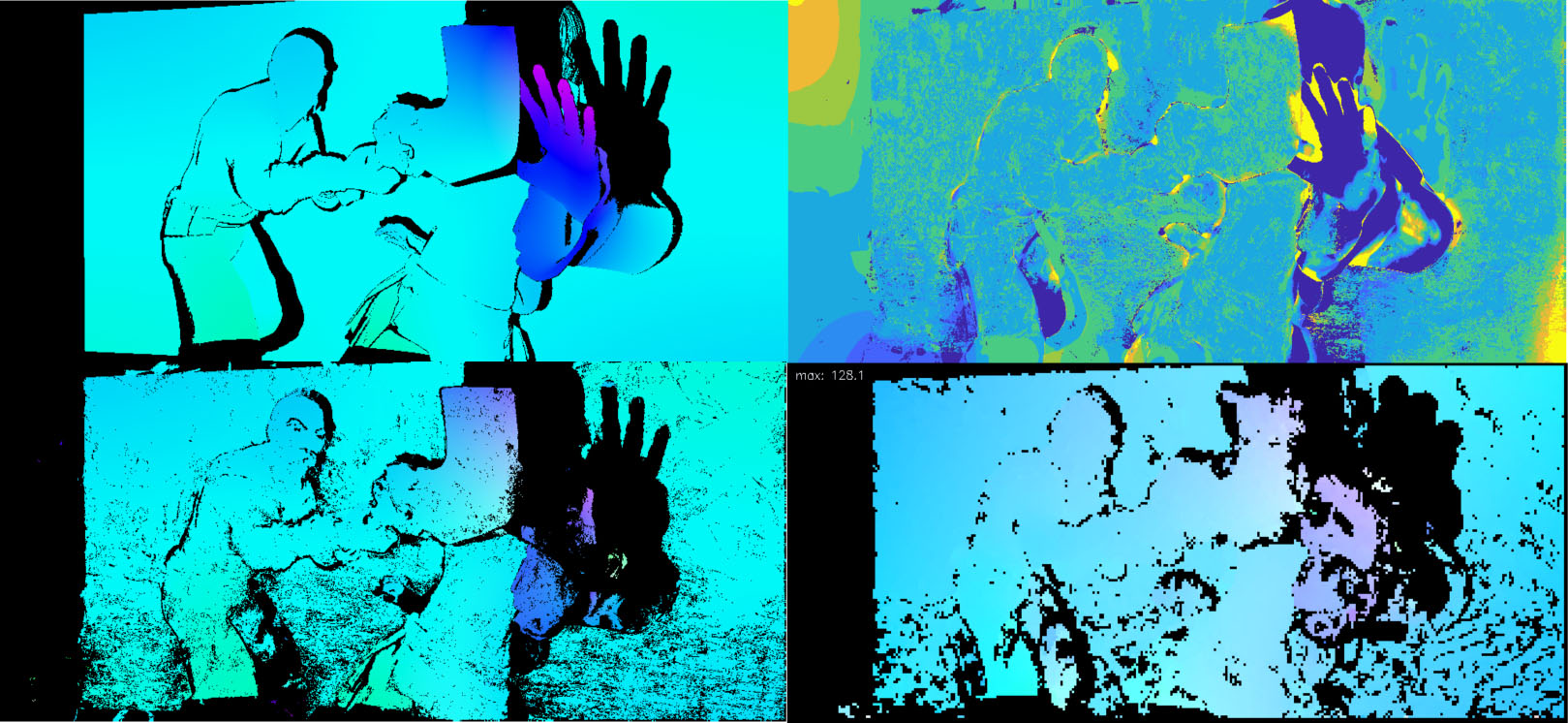}
\includegraphics[width=\linewidth, height=6cm]{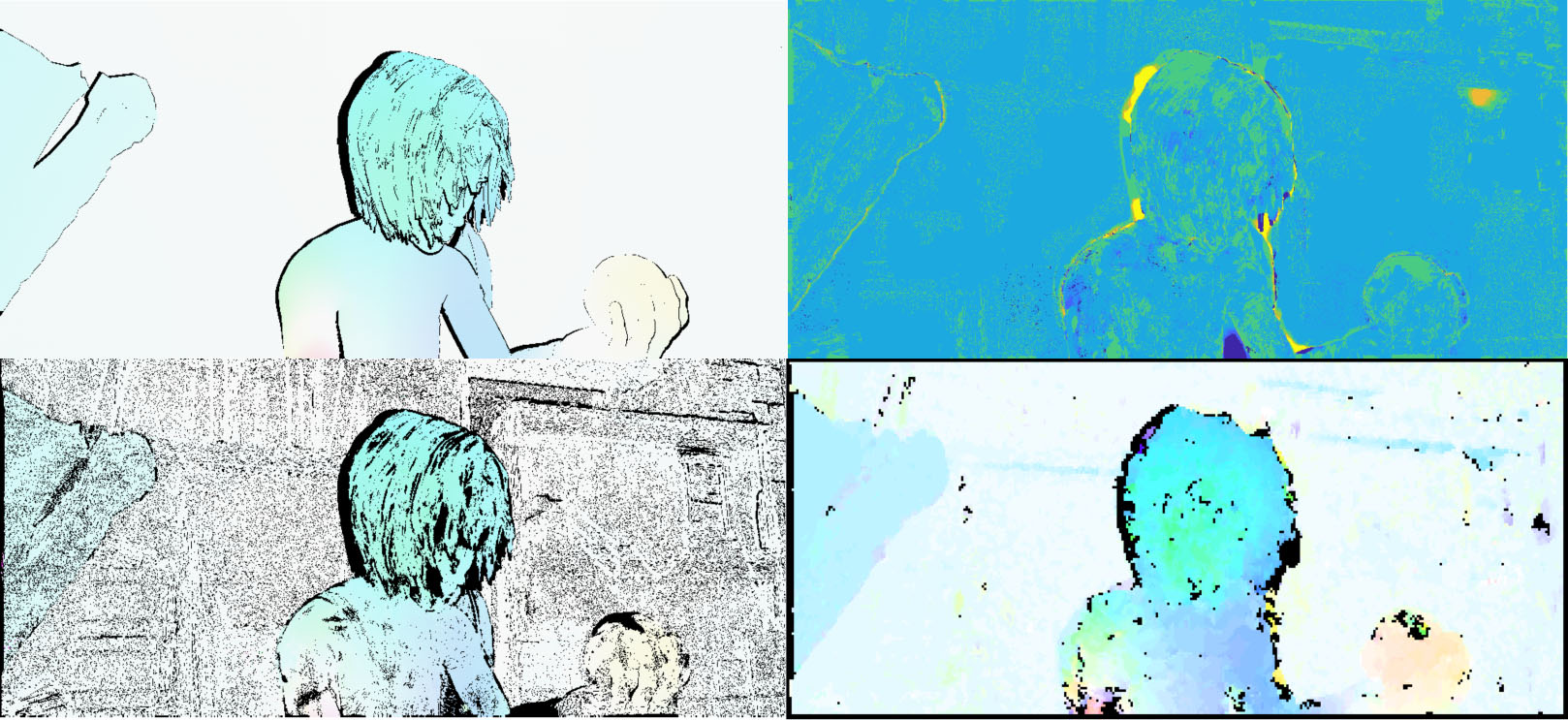}
\includegraphics[width=\linewidth, height=6cm]{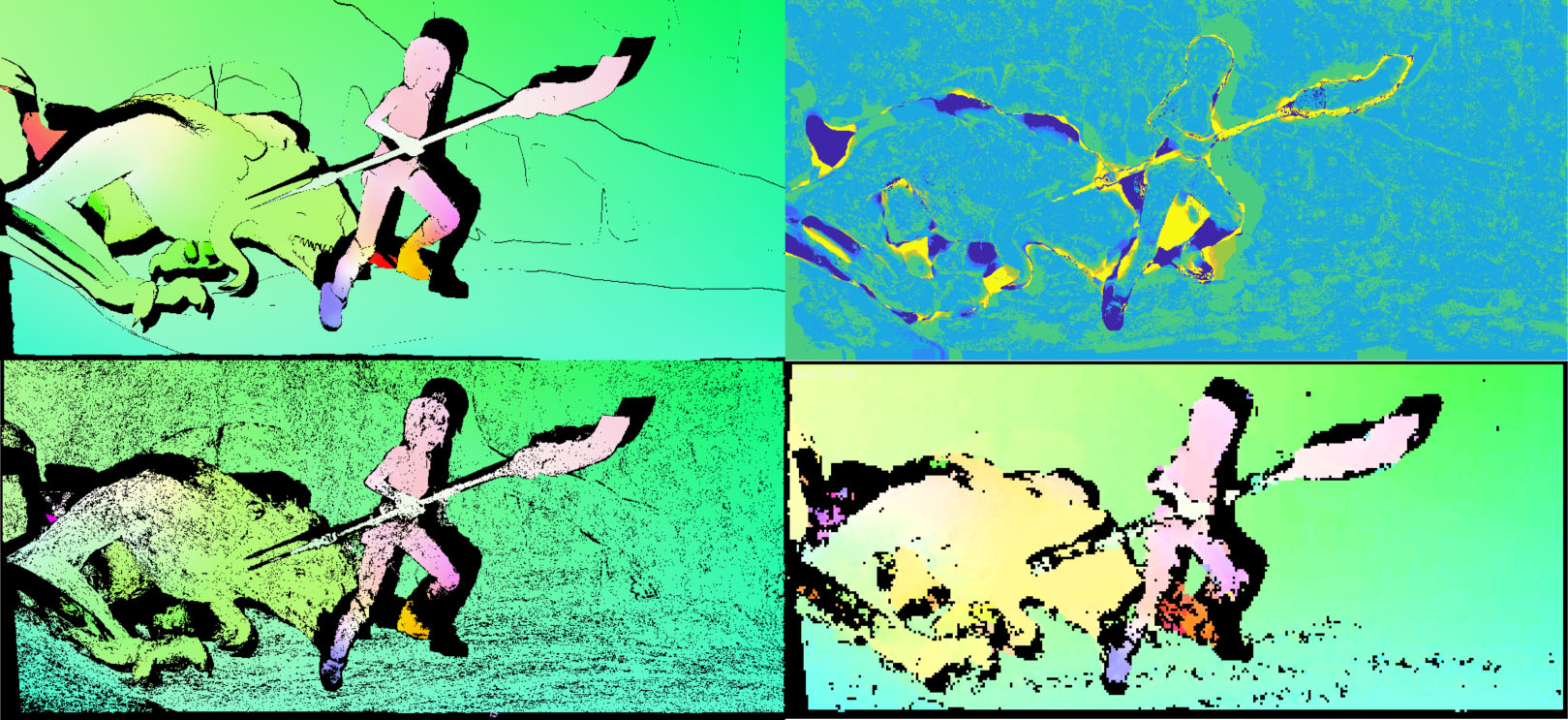}
\caption{EPE difference map and matches from MSGPM and CPM2 before EPIC interpolation. (for each image set) upper row left: Ground truth optical flow, upper row right:  EPE difference map between CPM2 and MSGPM, lower row left: MSGPM matched result, lower row right: CPM2 matched result}
\label{fig:epediff}
\end{figure}

\begin{figure}
\centering
\includegraphics[width=\linewidth, height=6cm]{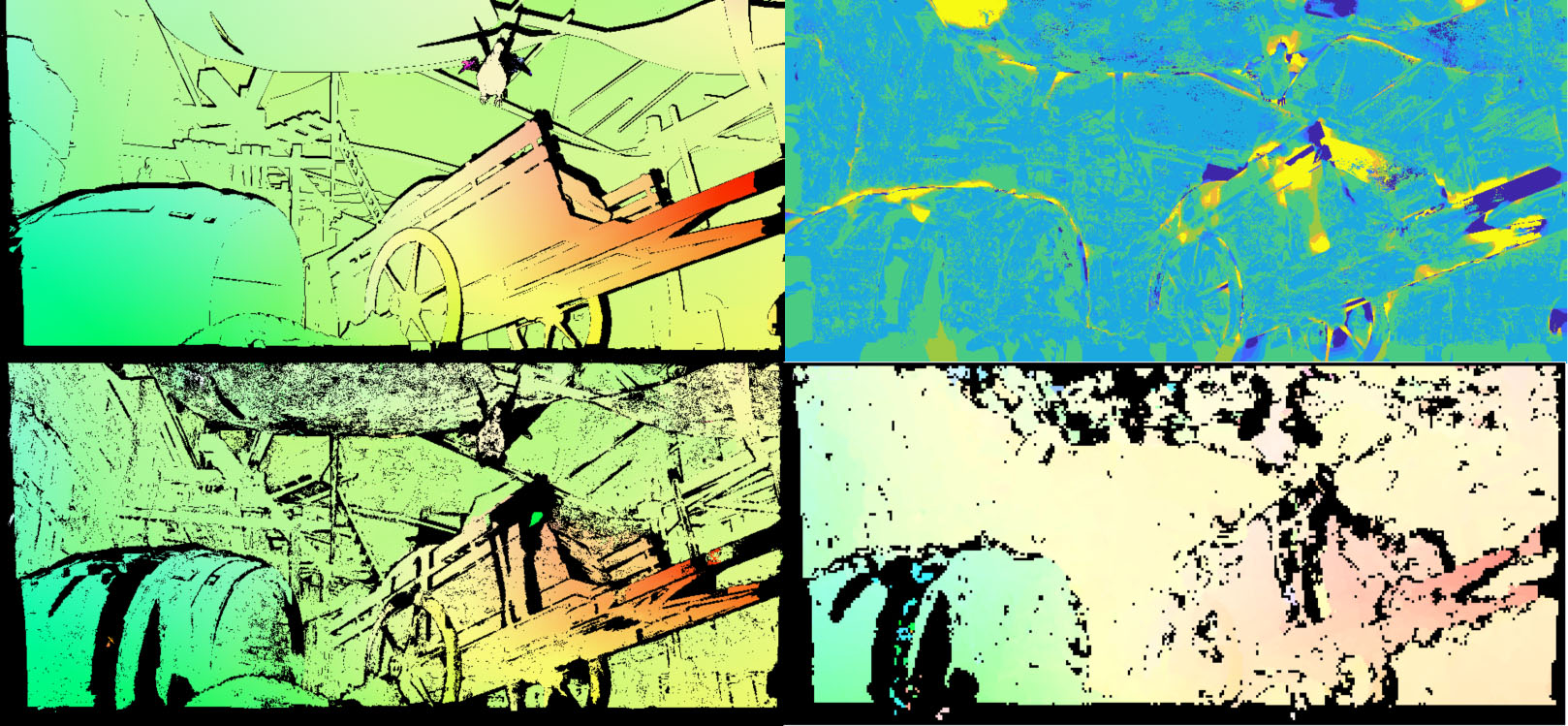}
\caption{EPE difference map and matches from MSGPM and CPM2 before EPIC interpolation. (for each image set) upper row left: Ground truth optical flow, upper row right:  EPE difference map between CPM2 and MSGPM, lower row left: MSGPM matched result, lower row right: CPM2 matched result}
\label{fig:epediff2}
\end{figure}

We found that MSGPM+Epic has superior mean EPE on the occluded regions of the
data set and slightly worse result on the overall mean EPE compared to the
other methods. To further analyze this, we compared MSGPM with CPM2 for which
code is available in the public domain.

We plot the EPE difference map ( CPM2 EPE - MSGPM EPE ) in
Fig.\ref{fig:epediff} and Fig.\ref{fig:epediff2} for example pairs of frames. The light blue region in the
difference map represents locations where CPM2 has a smaller EPE than MSGPM
within a one pixel difference bound. The yellow colored regions are where MSGPM
has a smaller EPE than CPM2.  As we can see in the figure, CPM2 has a slightly
smaller error in the large background area which contributes to its overall
smaller mean EPE.  We believe this effect comes from the difference in the flow
assignment strategy of the two methods. In CPM2, the best match is searched and
assigned to each pixel at a pixel by pixel basis. However, in MSGPM, the match
is assigned to all inlier pixels inside the smallest scale window with a single
dominant plane model.  Therefore, although MSGPM finds the most close plane
model to the local region, after assignment, it loses pixel-wise fine details
present in the true flow. And in this experiment, we used the smallest window
size as $41 \times 41$ which is quite a large size compared to CPM2's
pixel-wise assignment.  For the regions with non-rigid motion of objects, a
plane is difficult to be detected without using very small patch windows.
Therefore, MSGPM has less inliers in the region of non-rigid motion when
compared to CPM2. We expect this problem can be addressed by adding additional
finer scale levels with smaller patch window size to MSGPM.

\subsection{Thin Object Detection}

\begin{figure}
\centering
\includegraphics[width=\linewidth,height=6cm]{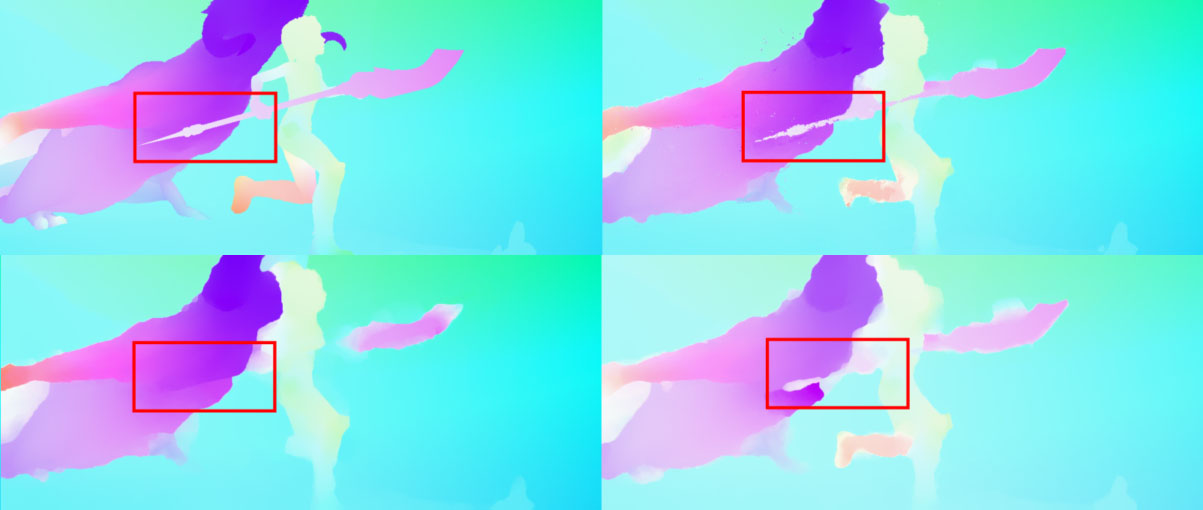}
\includegraphics[width=\linewidth,height=6cm]{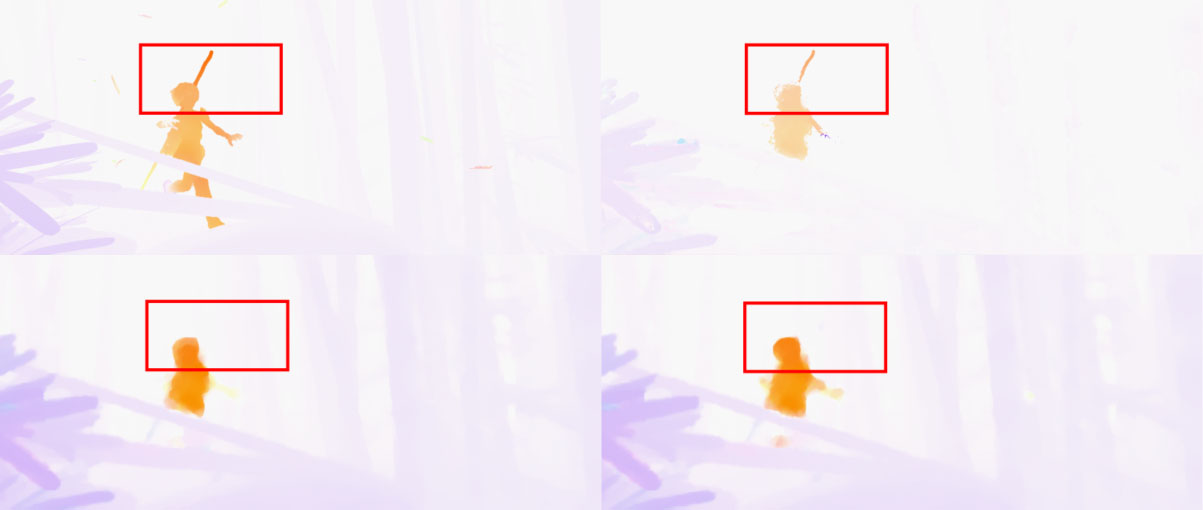}
\includegraphics[width=\linewidth,height=6cm]{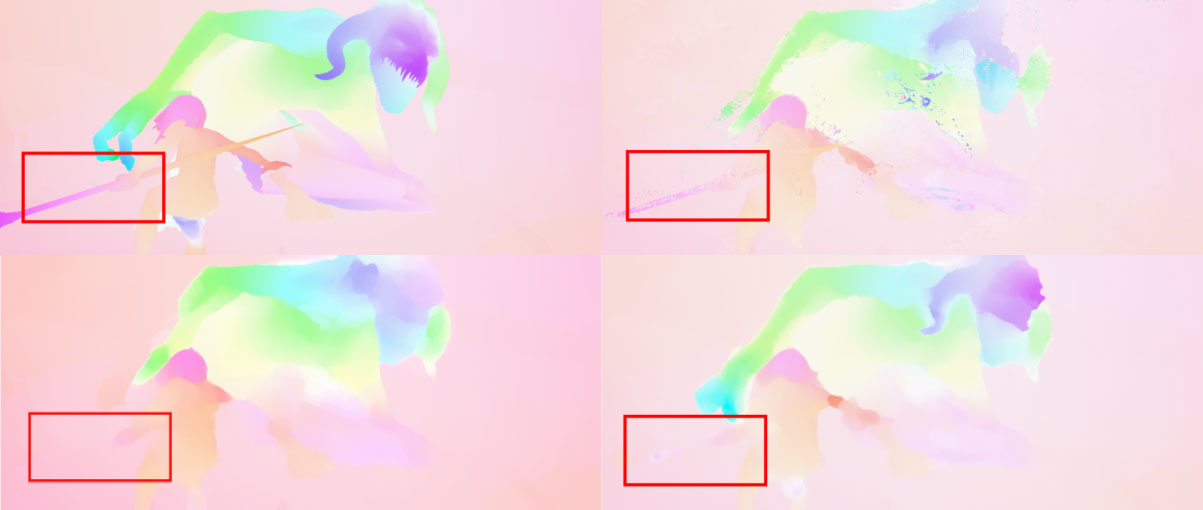}
\caption{Optical flow results. (for each image set) upper row left: Ground truth optical flow, upper row right:  MSGPM+Epic, lower row left: EPIC-flow, lower row right: CPM2-flow}
\label{spear}
\end{figure}

The detection of thin objects in a scene, particularly when they have large
movement, is a hard problem for dense match generation. The small area
encompassed by thin objects tend to be neglected during optical flow estimation
or the interpolation process. However, in the case of the MSGPM algorithm, such
thin objects were also localized and identified owing to the robust plane based
detection scheme. As demonstrated in Fig.\ref{spear}, thin structures such as
the spear in the scene is correctly identified by MSGPM in contrast to the
other techniques that failed to detect the spear in their results.

\begin{figure}
\centering
\includegraphics[width=\linewidth,height=6cm]{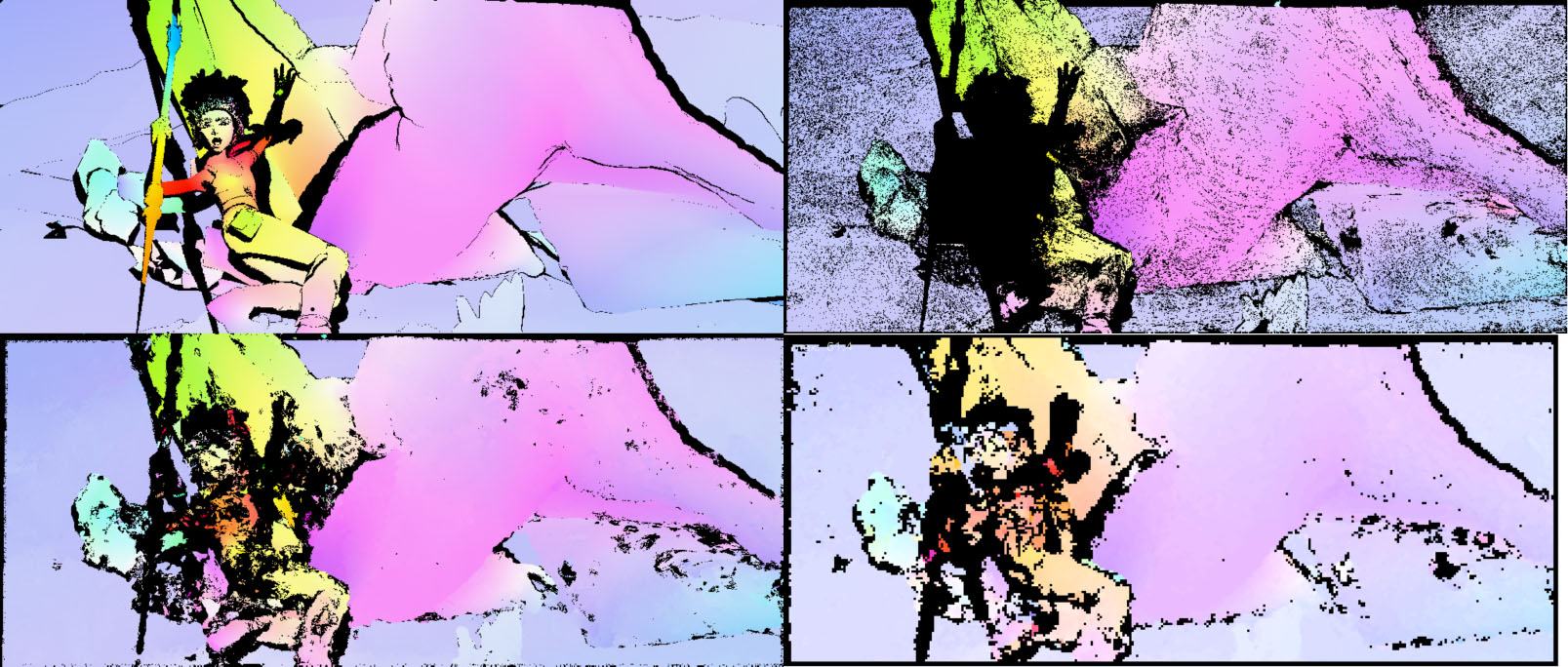}
\includegraphics[width=\linewidth,height=6cm]{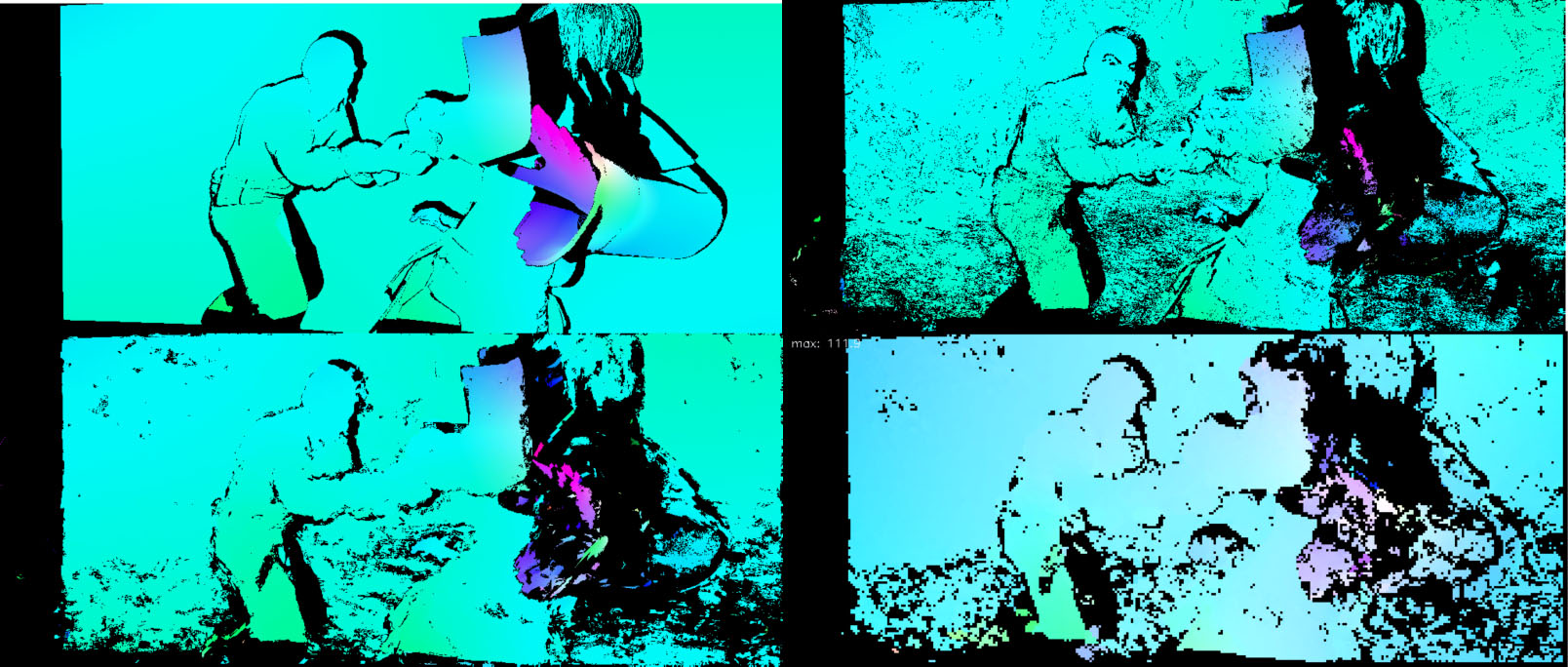}
\includegraphics[width=\linewidth,height=6cm]{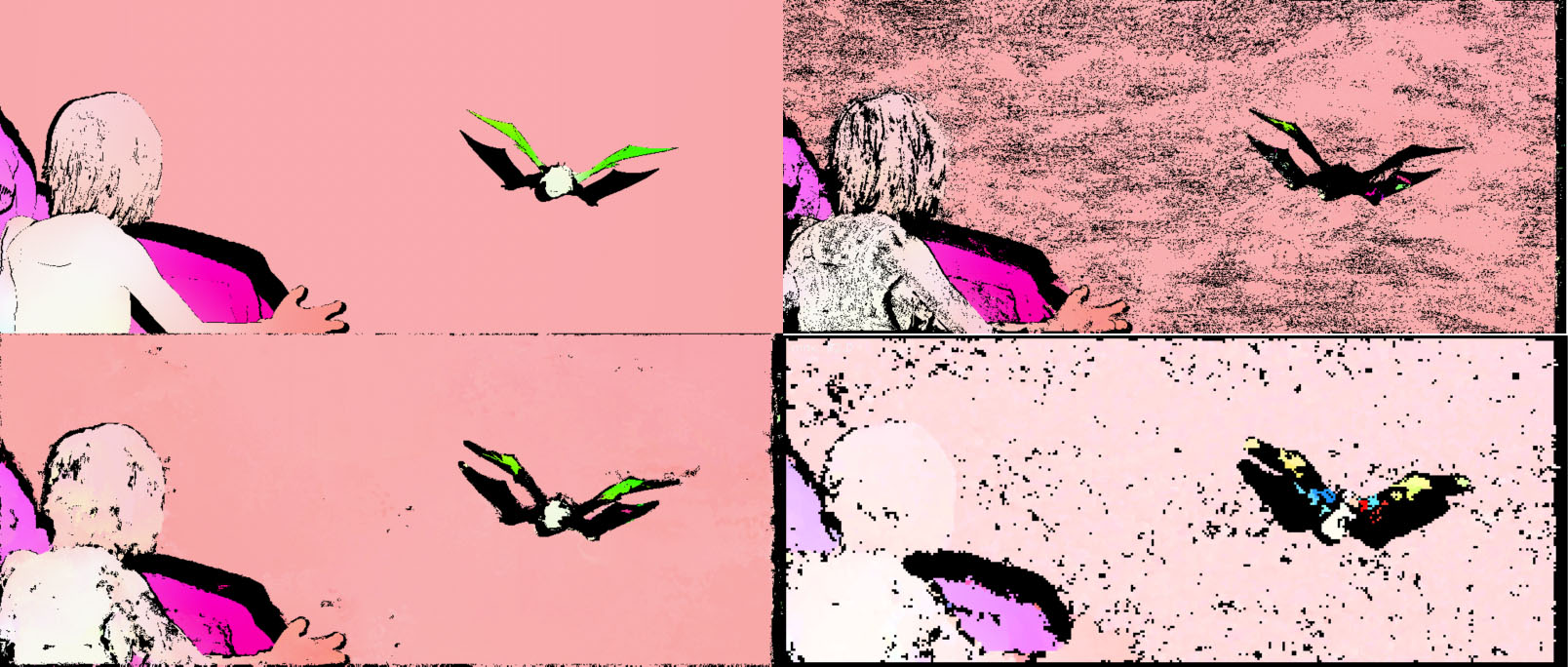}
\caption{Semi-dense correspondence before EPIC interpolation.(for each image set) upper row left: Ground truth optical flow, upper row right:  MSGPM match, lower row left: MSGPM match with histogram equalization, lower row right: CPM2 match before interpolation}
\label{histeqflow}
\end{figure}

\begin{figure}
\centering
\includegraphics[width=\linewidth,height=6cm]{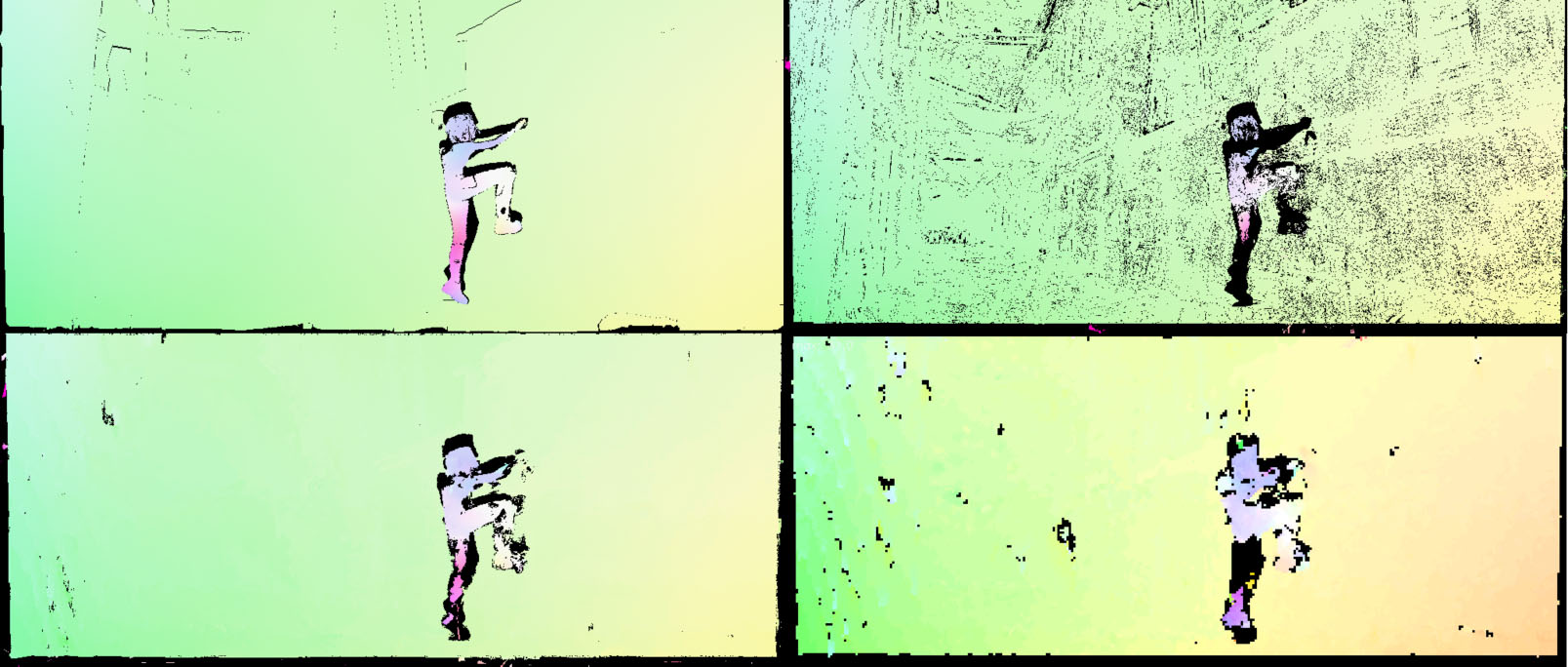}
\includegraphics[width=\linewidth,height=6cm]{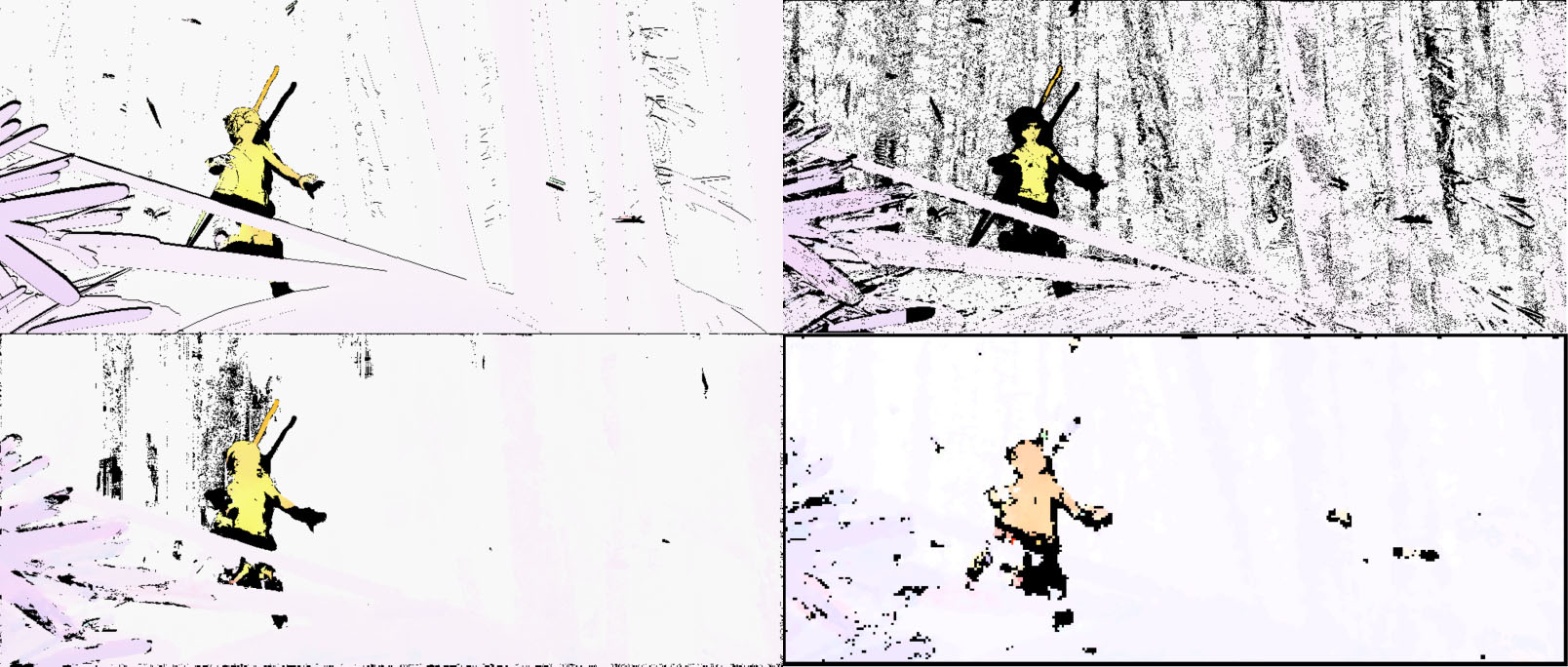}
\caption{Semi-dense correspondence before EPIC interpolation. (for each image set) upper row left: Ground truth optical flow, upper row right:  MSGPM match, lower row left: MSGPM match with histogram equalization, lower row right: CPM2 match before interpolation}
\label{histeqflow2}
\end{figure}

\subsection{Partial Experiments on Histogram Equalized Images}

In other experiments, we tested the case where we enhanced the illumination
robustness of the patch cost computation. Instead of using a more complicated
data term for patch cost computation, we simply applied histogram equalization
to the image pair in HSV format and ran the same MSGPM framework. As we can see
in Fig.\ref{histeqflow} and Fig.\ref{histeqflow2}, the simple histogram equalization can greatly improve
the quality of matches, without any modification of the MSGPM framework. The
histogram equalized MSGPM matches have more dense inliers and less outliers
compared to the MSGPM without histogram equalization and CPM2 generated
matches. This result shows that the MSGPM framework can be easily adapted to a
new enhanced data term for better correspondence generation.

\section{Conclusions}

We have presented a multi-scale generalized plane match based framework for
accurate optical flow and occlusion estimation.  Drawing from the insight that
patch based matching methods have an intrinsic limitation in the front-parallel
plane assumption, we derived a generalized plane based matching method that is
computationally tractable. From an initial NNF, we robustly detect dominant
plane models within a moving patch window and use it to compensate for plane
motion. Furthermore, from the obtained plane models we directly produce a
accurate optical flow.  This plane matching approach naturally reveals occluded
regions as those that do not have matching planes in the second image. Finally,
the technique is even able to identify thin structures that are rarely detected
by other state-of-the-art algorithms.  The evaluation on challenging datasets
shows that our approach is capable of providing accurate optical flow that is
comparable to other state-of-the-art algorithms in addition to precise
occlusion localization.

\section{Future Work}

The MSGPM is a powerful framework that can generate reliable matches from noisy
NNF, and furthermore, is highly flexible in the sense that it can easily be
adapted to a different kind of patch cost computation method.  We have shown
how this flexibility can further improve match quality in the experiments on
histogram equalized data sets. This motivates the possibility that  we can
further modify the technique to apply a new data term for illumination robust
matching and plane assignment, such as census transform or dense SIFT.
Furthermore, MSGPM is comprised of many operations that can be performed in
parallel.  Each patch-wise robust plane detection can be run independently at
the same time without affecting other plane detection operations in a single
level. If these operations are processed in parallel, it could save a great
deal of execution time.  Finally, we have used only 2 levels to obtain
results that are similar to the other state-of-the art algorithms. Therefore,
by increasing the number of levels, we can expect more accurate dense matching
results from MSGPM.  Finally, if we use a smaller window size for the finest
scale of MSGPM and add post-processing for the optimization of the fine details
of the flow in the plane assigned region, we can expect better optical flow
results as compared to other state-of-the-art optical flow algorithms.

\bibliographystyle{splncs}
\bibliography{reference}
\end{document}